# Vanishing Point Detection with Direct and Transposed Fast Hough Transform inside the neural network

*A. Sheshkus[4,6], A. Chirvonaya[2,6], D. Matveev[5,6], D. Nikolaev[1,6], V.L. Arlazarov[3,4]*
[1] *Institute for Information Transmission Problems (Kharkevich Institute) RAS, Moscow, Russia;*
[2] *National University of Science and Technology "MISIS";*
[3] *Moscow Institute for Physics and Technology, Moscow, Russia;*
[4] *Institute for Systems Analysis, Federal Research Center "Computer Science and Control" of Russian Academy of Sciences, Moscow, Russia;*
[5] *Lomonosov Moscow State University, Moscow, Russia;*
[6] *Smart Engines Service LLC, Moscow, Russia;*

*Abstract*

In this paper, we suggest a new neural network architecture for vanishing point detection in images. The key element is the use of the direct and transposed Fast Hough Transforms separated by convolutional layer blocks with standard activation functions. It allows us to get the answer in the coordinates of the input image at the output of the network and thus to calculate the coordinates of the vanishing point by simply selecting the maximum. Besides, it was proved that calculation of the transposed Fast Hough Transform can be performed using the direct one. The use of integral operators enables the neural network to rely on global rectilinear features in the image, and so it is ideal for detecting vanishing points. To demonstrate the effectiveness of the proposed architecture, we use a set of images from a DVR and show its superiority over existing methods. Note, in addition, that the proposed neural network architecture essentially repeats the process of direct and back projection used, for example, in computed tomography.

<u>Keywords:</u> Fast Hough Transform, vanishing points, deep learning, convolutional neural networks.

<u>Citation:</u> **Sheshkus A, Chirvonaya A, Matveev D, Nikolaev D, Arlazarov VL.** Vanishing Point Detection with Direct and Transposed Fast Hough Transform inside the neural network. Computer Optics 20XX; 4X(X): XXX-YYY. DOI: 10.18287/2412-6179-20XX-4X-X-XXX-YYY.

<u>Acknowledgements</u>: This work was supported by the Russian Foundation for Basic Research (projects 18-29-26027 and 17-29-03161).

### Introduction

The vanishing point (VP) is the intersection point of 2D projections of straight lines that are parallel in 3D space. In the field of computer vision, the problem of VP detection regularly arises in a great number of applications. It includes an analysis of both 2D and 3D scenes using images obtained by various types of cameras. In the 2D case, for example, we encounter all kinds of flat objects, such as documents with text (ID cards, bank cards, scanned pages, etc.). The current solution, in this case, begins with searching for the rectangle of the object [1] followed by straightening [2] for further recognition. The problem is that it is not always possible to find the rectangle because it may be beyond the edges of the image, merged with the background or obscured by other objects. There is an alternative group of methods that solve this problem using VP detection. In the case of a 3D scene, the detection of vanishing points is necessary to find objects, to assess their orientation or the orientation of the camera [3]. An example is shown in Figure 1. The vanishing point there (V in Figure) is the intersection point of the road edges. Similar images are used in this paper for VP detection.

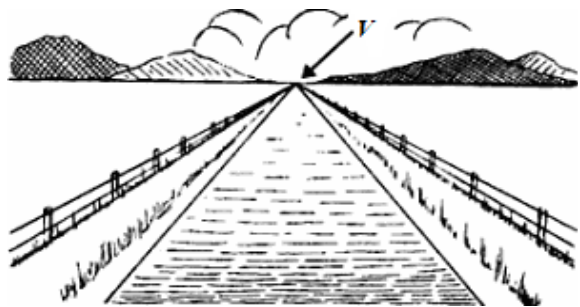

*Fig. 1. Vanishing point*

The classical approach to VP detection is presented in the work of Stephen Barnard [4]. The author uses a Gaussian sphere located in the optical center of the camera. Each point in the image corresponds to a point on the sphere, which is regarded as a radius vector. In this way, we can get mappings of infinitely distant points into a finite space and process them using conventional methods. In this case, the radius vector of an infinitely distant point will have the zero coordinate $z$. To find vanishing points, it is necessary to find the intersection points of all the lines in the image, and then to combine them into clusters. All the lines belonging to the same cluster represent a bunch of parallel straight lines in a certain perspective. For example, in [5], the authors propose to find the baselines of a text and the inclination angle of the characters using clusters of points on the Gaussian sphere. The main problem of the method is that the detection of straight lines is not so simple in images of natural scenes.

Another approach to VP detection is based on finding the intersection point of lines in the image. The following model is used to describe it: Let $P = p_i$, $i = 1...n$ be straight lines on flat images representing parallel lines in space. The line $p_i$ is described by a linear equation:

$$p_i = \{(x, y) \mid a_i x + b_i y = c_i\} \qquad (1)$$

Thus, the vanishing point $V(x, y)$ is an approximate solution to the system of linear equations 1 because there may not be an exact solution (when, for example, there are more than two straight lines and there is no unique intersection point). To find the VP using this model, we need to find straight lines in the image. A standard approach to finding them is to use the Hough Transform (HT). The paper [6] demonstrates various possible applications of such algorithms.

In [7], the author uses this algorithm to detect three vanishing points, but he notes that his method works only when applied to good synthetic data. The authors of [8] search for the VP successively applying two Hough transforms, but this method is very unstable to noise and the presence of outliers. A specific application of the Hough Transform for camera calibration is presented in [9]. The authors do not use information about camera parameters, but instead, transform the image of a chessboard because it contains a set of contrasting orthogonal lines that are easy to detect.

Recent papers show that artificial neural networks began to be used in VP detection. For example, [10] contains a solution to the problem of finding a vanishing point in images of road views using convolutional/fully connected neural network architectures with a large number of learning parameters (AlexNet, VGG). It also can be seen there that the method shows a low quality of work on images other than images from the training sample. Another application of convolutional neural networks in a similar problem is proposed in [11]. The authors trained the network to detect the horizon lines in the image. However, the use of these architectures for such tasks contains one serious problem: in the general case, the problem of the vanishing point detection cannot be solved only on local features, as it is done in fully convolutional networks, and the application of fully connected layers to the whole picture typically adds a huge number of learning parameters and significantly increases the amount of data required for training.

The idea of the convolutional network and FHT combination for VP detection have already been discussed [12]. The architecture suggested in that work contained two FHT layers interlaid by convolutional ones and had a fully-connected layer at the end. Such kind of architecture was used for proper comparison with the results stated in [10], where authors did not use FHT layers but ended the architecture with the fully-connected one. The problem of this layer is that poorness of train data makes it learn particular coordinates. However, removing the fully-connected layer from the previously suggested architecture will lead to the input connected with the output by the piecewise projective transform, whereas using the transposed FHT layer allows to get the output in the original image coordinates.

Note, that all the approaches [13] based on Hough transform use only its direct version. It means that finding VP location in the original image relying on the answer of such an algorithm is non-trivial. However, using it in pair with another transform converting a coordinate system into the initial state can make the algorithm more convenient to use. For example, an inverted Hough transform may be used, but the task of inversion is very complicated for that kind of transforms [14]. In [15] authors use the Moore-Penrose Pseudo Inverse because the matrix of the forward-projection operator cannot be inverted. Taking into consideration the presence of compound filtering between integral layers we can use transposed Hough transform instead. This transform is easy to construct, and in our work, we will prove that it is possible to compute it with the help of direct Fast Hough transform.

In this paper, we propose the architecture of a neural network, which is a combination of convolutional layers, of the Fast Hough Transform (FHT), and of a transposed FHT, which will enable the neural network to use not only local (as is the case with fully convolutional neural networks), but also global features. The idea of such a network architecture was based on the provided proof of the possibility to compute the transposed FHT using direct one, that allowed to exclude the explicit matrices multiplication. The interpretation of the network response will be reduced to a simple choice of the maximum owing to the use of the transposed HT. The effectiveness of the proposed approach will be demonstrated based on the task of the VP detection on road images taken from DVRs.

### 1. The proposed approach

The proposed method is very similar to the filtered back projection used in computer tomography [16]. It is based on the Radon Transform, discrete form of which is often referred as Hough transform.

The simplified idea of using a combination of the FHT and the transposed HT is shown in Figure 2. As an example, we took image 2*a* containing three lines with a common intersection point and a fourth line lying separately. Image 2*b* is the result of applying the FHT; it shows four blurry points, one for each straight line in the original image. The result of applying the transposed HT for a set of mostly horizontal lines to image 2*b* is shown in image 2*c*. The brightest point corresponds to the line on which the most points of the image 2*b* lie. Figures 2*a-c* are inverted for visibility.

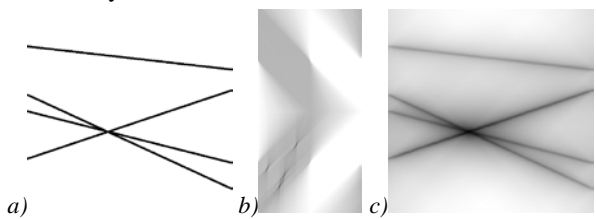

*a)* *b)* *c)*

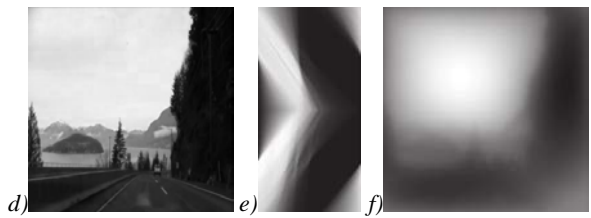

*Fig. 2. Hough Transform for vanishing point*

Here we did not include the results of the transformation along vertical lines because they did not provide any useful information in this case (there are only mostly horizontal lines in the image).

But applying the direct and transposed HT to natural images does not simplify the task of the vanishing point detection (see fig. 2*d*, 2*e*, 2*f*). An additional non-trivial image filtering is required in this case at each of the steps (as well as filtering the features of lines with certain inclination angles in the first case), which in our method will be done by blocks of convolutional layers. The paper [17] shows that the detection of straight lines by means of the HT with non-ideal data and the presence of overshoots implies the use of window operations. In addition, such a network will be able to rely not only on intensities along rectilinear objects but also on more complex nonlinear statistics. Although there are studies where the Hough transform is used in conjunction with neural networks, for example, as described in the papers [13,18], we failed to find any references to learning through it.

It is expected that the output of the last convolutional layer will produce an image containing bright points at positions related to vanishing points in the input image. There are several options for further actions. For example, the image can be covered with a grid of arbitrary size, and the softmax classifier can be trained to "recognize" the cell that contains the desired vanishing point, as it is done in [12]. Alternatively, we can construct an estimate of each point $(x, y)$ of the original image, which denotes the probability that it is a vanishing point. But in our case, the pixel with the maximum brightness value at the output of the last convolutional layer will correspond to the vanishing point in the original image.

## 2. Basic units

### Neural network layers

An artificial neural network is an information processing paradigm built from the model of biological neural network functioning. It was first introduced in 1943 by Warren McCulloch and Walter Pitts [19]. The network is based on a set of connected units called artificial neurons. The neurons are combined into layers of various types, which can perform different kinds of transformations of input data.

The neural networks with convolutional layers (convolutional neural networks) have been used since 1980 [20]; they have been developing rapidly since then and are currently one of the most popular and powerful image analysis tools. The input and output data of convolutional neural networks are images. A convolutional layer consists of a set of filters, each of which has its kernel. The filters are applied to different image areas spaced with a predetermined interval, for which reason convolutional layers are much less likely to be overfitted than fully connected ones, which ensures a better generalizing ability. The main drawback of convolutional neural networks is their high consumption of computational time resources, but numerous methods have already been developed to solve this problem from concurrent recognition on the GPU and CPU using fixed-point arithmetic to the tensor decomposition of filters.

### Fast Hough Transform

The Hough transform is a linear transform [20] that associates each straight line in the input image with a point in the output image. The result is the space $H \subset R^2$. The point $(s, \alpha) \in H$ contains the sum of the pixel intensities of the input image $I$ along a line $l$ where s is the distance from the line to the origin and $\alpha$ is the angle between the line and the positive direction of the abscissa axis on $I$. That is, $l(s,\alpha) = \{(x, y) \mid s = x\cos\alpha + y\sin\alpha\}$.

$$H(s, \alpha) = \sum_{(x,y) \in l(s,\alpha)} I(x, y). \qquad (2)$$

The computational complexity of the classical algorithm is $O(n^3)$, where $n$ is a linear size of the image. This paper uses the Fast Hough Transform [6] to save processing time. This version of the algorithm is faster due to the use of self-similar patterns for integration along straight lines with the calculation of partial sums. It is important to note that such recursive patterns approximate real lines in images with an accuracy sufficient for most problems [22, 23]. The computational complexity of the Fast Hough Transform is $O(log(n)n^2)$.

In this part we consider the FHT algorithm for mostly horizontal straight lines directed downward for the images of size $(2^p, 2^p)$, where $p \in \mathbb{N}$.

The FHT algorithm is defined in the paper [22]. Input and output of this algorithm are images that have the same size. The brightness of each pixel in the output image is the sum of the brightness of pixels belonging to one discrete line (dyadic pattern) in the input image. You can see all dyadic patterns of length 4 in figure 3. A pixel $Out(x_0, \theta)$ in the output image contains information about the pattern that have the slope $\theta$ pixels and starts at the pixel $I(x_0, 0)$. In the paper [22] author gives the analytic definition of patterns. This definition is equivalent to the recursive definition (that can be derived from the definition of the FHT algorithm). Let us define the function of indentation of patterns' pixels: if $n = 2^p$ is a length of the pattern, $t = \overline{0, n-1}$ is the indentation of the whole pattern, $x = \overline{0, n-1}$ is the number of the pixel position, $t = \sum_{r=0}^{p-1} t_r 2^r$ and $x = \sum_{i=0}^{p-1} x_i 2^i$ are decompositions of $x$ and $t$ by powers of 2, then the indentation equals $H(x,t) = \sum_{r=0}^{p-1} t_r [\frac{2^r x}{2^p - 1}]$.

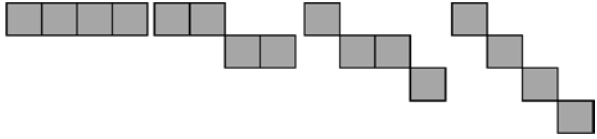

*Fig. 3. Patterns of length 4*

### Transposed Hough Transform

In this paragraph, we will prove that for calculating transposed Fast Hough Transform we just need to flip vertically input and output image of FHT algorithm.

The set of all images with shape $(n, n)$ can also be considered as a vector space $X$. Let us enumerate image pixels row-wise: $I_{i*n+j} := \tilde{I}_{ij}, i, j = \overline{1,n}, \tilde{I} \in X$ and assign basis vector $x_m$ to pixel $I_m$ for each $m$. Define the inner product in this space by the dot product. The brightness of each pixel in the output image is the linear combination of brightness of pixels in the input image and the output image $\in X$, hence the action of the FHT algorithm can be considered as the action of the linear operator $\mathcal{A}: X \to X$. Denote by $A$ matrix of $\mathcal{A}$. $A$ will be also called the FHT matrix. The Gram matrix of $X$ is the identity matrix, hence the matrix of $\mathcal{A}^*$ (transpose of $\mathcal{A}$) is $A^T$.

**Lemma 1.** $H(x, t) = H(t, x)$ for fixed $n$. In other words, a matrix that consists of the patterns indentations is symmetric.

Such a matrix for the patterns of length 4 is presented in figure 4.

| 0 | 0 | 0 | 0 |
|---|---|---|---|
| 0 | 0 | 1 | 1 |
| 0 | 1 | 1 | 2 |
| 0 | 1 | 2 | 3 |

*Fig. 4. Matrix of patterns indentations*

Proof of this lemma was given in paper [24], but it was based on another definition of the pattern. In the paper [23] the author uses the same definition as was given earlier, but he uses this statement without proof.

*Proof.* Let us replace $x$ by its decomposition by powers of 2 and then replace the fraction inside the sign of rounding by the sum of fractions:

$$H(x,t) = \sum_{r=0}^{p-1} t_r [\frac{2^r x}{2^p - 1}] = \sum_{r=0}^{p-1} t_r [\sum_{i=0}^{p-1} \frac{2^r x_i 2^i}{2^p - 1}]. \quad (3)$$

Note that $\forall y \in \mathbb{R} \ \frac{2^y}{2^p - 1} = 2^{y-p} + \frac{2^{y-p}}{2^p - 1}$, therefore $\forall r = \overline{0, p-1}$:

$$\{\{\frac{x_i 2^r 2^i}{2^p - 1}\}, i = \overline{0, p-1}\} = \{\{\frac{x_i 2^i}{2^p - 1}\}, i = \overline{0, p-1}\} =: S \quad (4)$$

Let us note two important properties of $S$: only one element in $S$ can be bigger than 1/2: $\frac{x_{p-1} 2^{p-1}}{2^p - 1}$, and also the sum of all elements in $S$ is not bigger than 1:

$$\sum_{i=0}^{p-1} \frac{x_i 2^i}{2^p - 1} \le \sum_{i=0}^{p-1} \frac{2^i}{2^p - 1} = \frac{1}{2^p - 1}(\sum_{i=0}^{p-1} 2^i) = 1. \quad (5)$$

Now let us rewrite (3):

$$[\sum_{i=0}^{p-1} \frac{x_i 2^r 2^i}{2^p - 1}] = [\sum_{i=0}^{p-1} (\lfloor \frac{x_i 2^r 2^i}{2^p - 1} \rfloor + \{\frac{x_i 2^r 2^i}{2^p - 1}\})] =$$
$$= \sum_{i=0}^{p-1} \lfloor \frac{x_i 2^r 2^i}{2^p - 1} \rfloor + [\sum_{i=0}^{p-1} \{\frac{x_i 2^r 2^i}{2^p - 1}\}]. \quad (6)$$

Note that if $\{y\} < \frac{1}{2}$, therefore $\lfloor y \rfloor = [y]$. From this and two properties of $S$ we can conclude that

$$\sum_{i=0}^{p-1} \lfloor \frac{x_i 2^r 2^i}{2^p - 1} \rfloor = \sum_{i=0}^{p-1} [\frac{x_i 2^r 2^i}{2^p - 1}] - x_{p-1-r}. \quad (7)$$

and

$$\sum_{i=0}^{p-1} \{\frac{x_i 2^r 2^i}{2^p - 1}\} = x_{p-1-r} \quad (8)$$

Finally combine results (3), (6), (7) and (8):

$$H(x,t) = \sum_{r=0}^{p-1} t_r [\sum_{i=0}^{p-1} \frac{2^r x_i 2^i}{2^p - 1}] \overset{6}{=}$$
$$\overset{6}{=} \sum_{r=0}^{p-1} t_r (\sum_{i=0}^{p-1} \lfloor \frac{x_i 2^r 2^i}{2^p - 1} \rfloor + [\sum_{i=0}^{p-1} \{\frac{x_i 2^r 2^i}{2^p - 1}\}]) \overset{7,8}{=}$$
$$\overset{7,8}{=} \sum_{r=0}^{p-1} t_r (\sum_{i=0}^{p-1} [\frac{x_i 2^r 2^i}{2^p - 1}]) =$$
$$= \sum_{r=0}^{p-1} \sum_{i=0}^{p-1} t_r x_i [\frac{2^r 2^i}{2^p - 1}]) = H(t, x). \quad (9)$$

◄

Let us consider the FHT matrix as a block matrix **B** with all blocks having a size $(n, n)$ (see figure 5). Blocks $\mathbf{B}_{i,j}$ will be also denoted as $(i, j)$-blocks.

$$A = \begin{pmatrix} \mathbf{B}_{1,1} & \mathbf{B}_{1,2} & \cdots & \mathbf{B}_{1,n} \\ \mathbf{B}_{2,1} & \mathbf{B}_{2,2} & \cdots & \mathbf{B}_{2,n} \\ \vdots & \vdots & \ddots & \vdots \\ \mathbf{B}_{n,1} & \mathbf{B}_{n,2} & \cdots & \mathbf{B}_{n,n} \end{pmatrix}$$

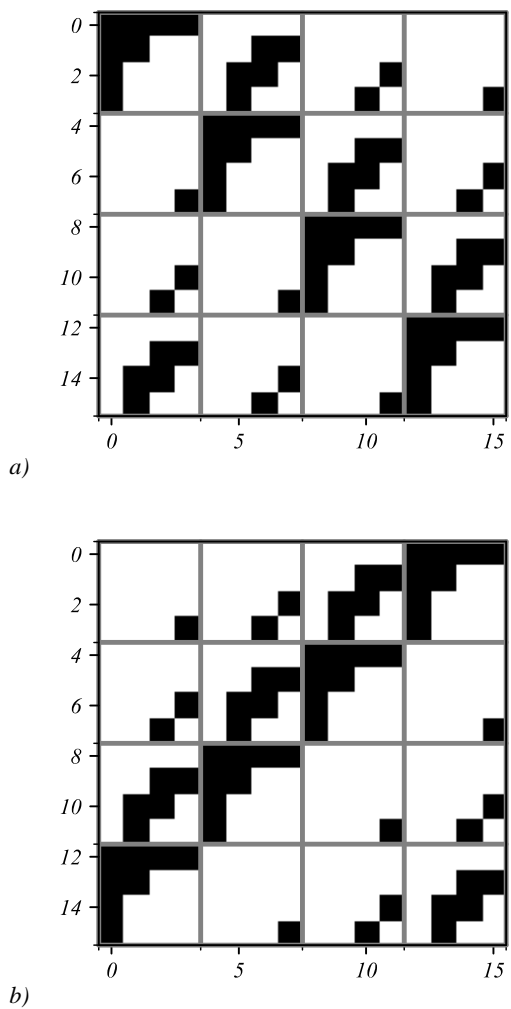

*Fig. 5. Compare blocks of matrices with equally numbered areas on figure 4: a) FHT matrix; b) FHT matrix if we will rotate input image before algorithm (AC)*

**Lemma 2.** Blocks are symmetrical.

*Proof.* $A_{m,q}$ is the projection of the input pixel $q$ on the output pixel $m$, therefore $\mathbf{B}_{i,j}$ is the projection of the input row of pixels $\{(j-1)*n+k, k=\overline{1,n}\}$ on the output row of pixels $\{(i-1)*n+k, k=\overline{1,n}\}$. Therefore

$$(\mathbf{B}_{i,j})_{k,l} = \begin{cases} 1, & \text{if } H(k,l) = n-i+j(\bmod\ n); \\ 0, & \text{else}, \end{cases}$$

where $H$ is previously defined indentation function. From lemma 1 $H(k,l) = H(l,k)$, therefore, $B_{i,j}$ is symmetrical. ◄

**Lemma 3.** $\mathbf{B}_{i,j} = \mathbf{B}_{i+k-1(\bmod\ n)+1, j+k-1(\bmod\ n)+1}$ for all $i, j = \overline{1,n}, k = \overline{0,n-1}$.

*Proof.* The FHT algorithm input image rows rotation will rotate the output image rows. The FHT matrix columns are the output images of $X$ basis vectors. Hence columns $x_i$ and $x_{i+n}$ are rotated cyclically by $n$ pixels, then $A_{i,j} = A_{i+n, j+n}$. Therefore, elements of $\mathbf{B}$ in one diagonal are equal. ◄

Denote $n \times n$ blocks of matrix $AC$ by $\tilde{\mathbf{B}}_{ij}$ (see figure 5b) and blocks of matrix $CAC$ by $\hat{\mathbf{B}}_{ij}$.

**Lemma 4.** $AC$ is symmetric.

*Proof.* Right multiplication of matrix $A$ by matrix $C$ permutes columns of the matrix $A$. Even more from the $C$ definition we can conclude that this multiplication permutes columns of the matrix $\mathbf{B}$, i.e. it does not change the disposition of the $A$ columns inside a group $\{(i-1)*n+k, k=1,n\}, i=1,n$. $\mathbf{B}_{i,j} = \tilde{\mathbf{B}}_{n+1-i,j}$, $\tilde{\mathbf{B}}_{j,n+1-i} = \mathbf{B}_{n+1-j, n+1-i} = \mathbf{B}_{(i,j)+(1,1)\times(n+1-i-j)}$. From lemma 3 $\mathbf{B}_{i,j} = \mathbf{B}_{(i,j)+(1,1)\times(n+1-i-j)}$, therefore, $\tilde{\mathbf{B}}$ is symmetric. From lemma 2 $\mathbf{B}_{i,j}$ are symmetric and thus $AC$ is also symmetric. ◄

**Theorem 1.** Let $\mathcal{C}$ be the operator that flips an image: $\mathcal{C}e_{i*n+j} = e_{(n+1-i)*n+j}$, where $i,j = \overline{1,n}$. Then $\mathcal{C}\mathcal{A}\mathcal{C} = \mathcal{A}^*$ (or $CAC = A^T$, where $C$ is a matrix of $\mathcal{C}$).

Note that $C = C^{-1}$, therefore $A^T$ is also a matrix of $\mathcal{A}$ in basis $<\mathcal{C}x_i>$. We will prove the theorem using lemmas 1, 2, 3, 4.

*Proof.* $\mathbf{B}_{i,j} = \tilde{\mathbf{B}}_{n+1-i,j}$. From lemma 4 $AC$ is symmetric therefore $\tilde{\mathbf{B}}_{n+1-i,j} = \tilde{\mathbf{B}}_{j,n+1-i}$. Left multiplication of matrix $AC$ by matrix $C$ permutes rows of matrix $AC$ and matrix $\tilde{\mathbf{B}}$. Therefore $\mathbf{B}_{i,j} = \tilde{\mathbf{B}}_{j,n+1-i} = \hat{\mathbf{B}}_{j,n+1-(n+1-i)} = \hat{\mathbf{B}}_{j,i}$. From lemma 2 all $\mathbf{B}_{i,j}$ are symmetric, therefore, $CAC = A^T$. ◄

*The Hough transform as a neural network layer*

To train a neural network to detect lines and vanishing points, it is necessary to calculate the FHT inside the neural network. Our first implementation was based on the idea that both the classical and fast Hough transform can be calculated using an untrainable fully connected layer with pre-calculated weights. This approach makes it possible to use standard training tools without any changes. The main disadvantage of this method is that it requires $O(n^2)$ memory, where n is the length of the input vector. Another problem is that the weight matrix contains many zero values, which indicates unnecessary time costs.

For this reason, the next step was the implementation of a new FHT layer, which simply calculates the transformation inside the neural network. Since learning layers are located before the FHT layer, it was necessary to implement the backpropagation of the gradient for it. We used the FHT for forward propagation and equation (10) for backpropagation, which corresponds to the transposed

Hough Transform. The equation was obtained numerically.

$$I_H(x,y) = \sum_{(s,\alpha) \in l(x,y)} H(s,\alpha), \quad (10)$$

where $I_H$ is the transposed Hough transform and $l_{xy} = \{(s,\alpha) \mid s = x\cos\alpha + y\sin\alpha\}$ is the set of Hough-parameters of all lines passing through the point $(x,y)$ in the space of the image. According to theorem 1 (10) can be computed using FHT algorithm in $O(\log(n)n^2)$ operations.

### 3. Experiments

We used data from [10] for a numerical experiment. This dataset consists of frames of records from DVRs of cars, buses, or trucks found on YouTube and made in traveling around America. The frame size was 300×300 pixels. There were 9972 images in total, which were randomly divided into a training sample (8974 images) and a test sample (998 images). Examples of the images are shown in Figure 6.

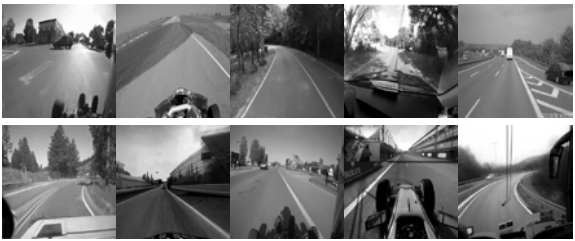

*Fig. 6. Images examples*

To estimate the error, we used the method also proposed by the author of the dataset. Each image was covered with 10×10, 20×20, and 30×30 grids, and the answer was considered correct if the resultant vanishing point fell into a cell with the correct answer. At first, we tried to use the proposed AlexNet architecture but later realized that the declared quality can be achieved with a simpler architecture of the neural network. The convolutional network architecture consisted of four convolutional layers and one fully connected layer. And the input images had to be resized to 227×227 pixels as required by the original architecture. The main difference between our base architecture and AlexNet is that it has a smaller number of convolution kernels and just one fully connected layer.

The detailed description of the basic architecture is presented in Table 1. The network with a new architecture was trained after training the core network. The innovation was training through the layers of the direct and transposed FHT, which we added between the layers of convolutions. Moreover, we removed the fully connected layer and searched instead for the pixel with the maximum brightness value in the output image because the image remained in the original coordinates after the application of the direct and transposed FHT. The description of the proposed architecture is presented in Table 2. Padding layers were used before applying the FHT and transposed HT to compensate for the reduction in image size due to the use of convolutional layers without padding. Thus, just a "frame" of zero values around the picture was created in these layers.

*Table 1. Base architecture*

| # | Type | Parameters | Activation function |
|---|------|------------|---------------------|
| 1 | Convolutional | 32 filters 11×11, no padding, stride 4×4 | relu |
| 2 | Convolutional | 32 filters 5×5, padding 2×2, stride 1×1 | relu |
| 3 | Convolutional | 32 filters 3×3, padding 1×1, stride 1×1 | relu |
| 4 | Convolutional | 32 filters 3×3, padding 1×1, stride 1×1 | relu |
| 5 | Fully-connected | | - |
| 6 | Softmax | | - |

*Table 2. Improved architecture*

| # | Type | Parameters | Activation function |
|---|------|------------|---------------------|
| 1 | Convolutional | 12 filters 5×5, no padding, stride 1×1 | tanh |
| 2 | Convolutional | 12 filters 5×5, no padding, stride 3×3 | tanh |
| 3 | Convolutional | 12 filters 3×3, no padding, stride 1×1 | tanh |
| 4 | Convolutional | 12 filters 3×3, no padding, stride 1×1 | tanh |
| 5 | Padding | 4×4 | - |
| 6 | FHT | | rf[3,1] |
| 7 | Convolutional | 12 filters 5×5, no padding, stride 1×1 | tanh |
| 8 | Convolutional | 12 filters 5×5, no padding, stride 1×1 | tanh |
| 9 | Convolutional | 12 filters 5×5, no padding, stride 1×1 | tanh |
| 10 | Padding | 6×6 | - |
| 11 | Transposed HT | | rf[3,1] |
| 12 | Convolutional | 12 filters 5×5, no padding, stride 1×1 | tanh |
| 13 | Convolutional | 12 filters 5×5, no padding, stride 1×1 | tanh |
| 14 | Convolutional | 12 filters 5×5, no padding, stride 1×1 | rf[2,1] |

The total number of learning coefficients in the basic architecture was 170967, 516576, and 1092576 for grids of 10×10, 20×20, and 30×30, respectively, while the new network had 25309 learning coefficients regardless of the grid size. It should be noted that we used a heavily modified version of cuda-convnet for training the networks [25]. In the proposed architecture, a hyperbolic tangent was chosen as an activation function for convolutional layers and the function (11) for layers following the direct and transposed FHT.

$$rf[a,b] = \frac{x^a}{b+|x^a|} \qquad (11)$$

It was found empirically that this kind of function qualitatively improves the network convergence. We assume that the presence of an inflection point at zero plays the role of an amplifier of local extrema on Hough images, but this issue requires additional research.

### 4. Results

As a result of this study, an improved algorithm of training the convolutional neural network through the Hough transform layers was implemented. Although this type of layer can be expressed with fully connected and theoretically there is no need to invent anything, the matrix of this layer will consist of $n^2$ units, where is the length of the input vector. For example, the weight matrix will have $4*10^{10}$ units for an image of 100×100×20 in size (relatively small for a convolutional neural network). The approach thus implemented makes it possible to calculate direct and back passages through the HT layers without using this type of matrix. Moreover, it was proved that it is possible to calculate transposed FHT without explicit matrices multiplication using direct FHT. It means that the computational complexity of the proposed transposed FHT algorithm is $O(log(n)n^2)$.

Figure 7 shows examples of the operation of the proposed algorithm and visualizes the purpose of the FHT layer. Since the intermediate outputs of the network are multichannel, we selected for illustrations the channels that we thought to be the most illustrative. 7*a* shows the original image, 7*b* is the image obtained after the first block of convolutions, 7*c* is the output of the fast Hough transform layer, 7*d* is the output of the FHT after the second block of convolutions, 7*e* is the output of the layer of the transposed FHT, and 7*f* is the image obtained at the output of the network - a bright spot at the assumed location of the vanishing point. It can be seen that by using the proposed method it is possible to train convolutional layers to select suitable straight lines and boundaries and to solve the problem after that.

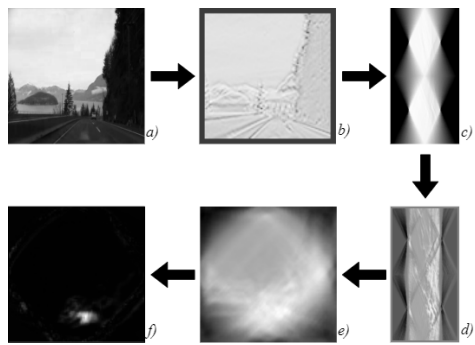

*Fig. 7. Examples from the neural network: a) Input image; b) Image after the first convolutions block; c) FHT layer output; d) Transposed FHT input; e) Transposed FHT output; f) Neural network output*

Table 3 presents error values for basic and proposed methods for various net sizes and different numbers of point alternatives. Our method demonstrated a higher detection accuracy. Moreover, it weakly depends on the size of the net due to the absence of a fully connected layer at the end.

*Table 3. Basic and suggested methods error values for different grid sizes*

| Grid size | Base – top 1, % | Base – top 5, % | Ours – top 1, % | Ours – top 5, % |
|---|---|---|---|---|
| 10×10 | 31.7 | 5.2 | **1.5** | **0.7** |
| 20×20 | 44.5 | 16.1 | **5.4** | **2.3** |
| 30×30 | 52.9 | 25.6 | **6.2** | **3.1** |

To show the tolerance of the proposed method to different image corruption we made an additional experiment. We took the suggested network and modified test data so that the rectangles of different sizes with centers at the positions of vanishing points became blurry. Examples of the corrupted data are depicted in figure 8.

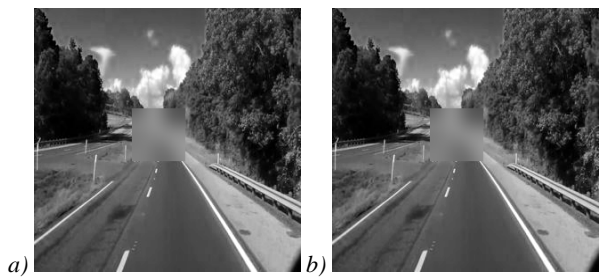

*Fig.8. Examples of defferent-sized images corruption: a)100x100; b)60x60*

The results of the second experiment where we used corrupted data are presented in figure 9. Graphs show error dependence on the size of a blurred area for different grid sizes. It can be noticed that proposed method does not require the presence of the vanishing point in the image to locate it. The reason for it is that FHT layers rely on global rectilinear features in the image. This fact makes it possible to broaden the class of problems that can be solved using this method.

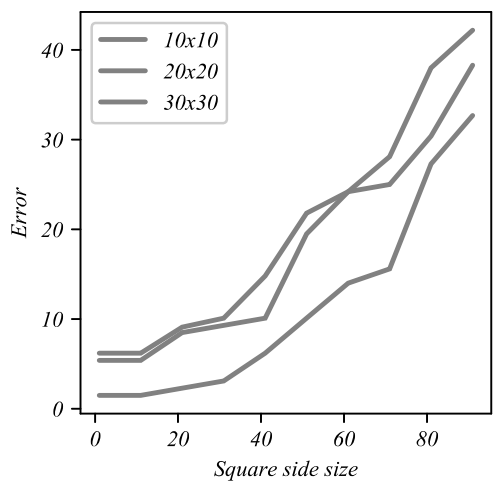

*a)*

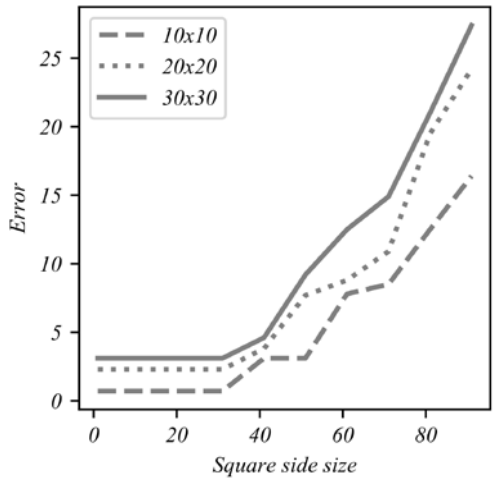

*b)*

*Fig. 9. Suggested method errors for grids of sizes 10x10, 20x20 and 30x30: a) Top 1 point; b) Top 5 points*

According to the graph, the error of the variant using top 5 points remains the same to a certain size of blurred area for every grid size. This points to the fact that partial image corruption does not affect the result. The absence of a vanishing point in the image is quite conceivable, for example, in a road scene it can be obstructed by a car. Thus, according to experiment results suggested method is applicable enough to this problem because truly uses global features rather than local ones. Consequently, none of the usual convolutional networks with the finite receptive field can approximate the network with the proposed architecture. The experiment demonstrates that the suggested network does not find a local intersection point itself but relies on lines segments around and finds their intersection point.

It is important to note that the FHT layer has no learning coefficients, so it does not complicate the network architecture in terms of the number of weights. All operations on this layer are predetermined and do not change in the process of training.

Finally, it is necessary to estimate the contribution of the FHT layer to the computational complexity of the algorithm. The number of the required operations is about $cs^2 log(s)$, where $c$ is the number of image channels and $s$ is the linear image size. The convolutional layer needs about $cs^2 f^2 m$ operations, where f is the linear size of the filter and m is the number of filters. The ratio of the computational complexities of the convolutional and FHT layers is $\frac{f^2 m}{\log(s)}$. Hence, the contribution is small and practically negligible in the case of deep convolutional networks.

### *Conclusion*

In this paper, we suggest a new neural network architecture based on using the direct and transposed Fast Hough Transforms. The theorem about the possibility to calculate transposed FHT without explicit matrices multiplication and with the same asymptotical complexity as the direct FHT was stated and proved. This theorem provided the basis for the proposed architecture. Its effectiveness was demonstrated on the task of vanishing point detection in road images. In this case, the main advantage of the proposed method is that it avoids attempts to detect the vanishing point in some position in favor of finding suitable elements in the input image using convolution filters and constructing the resultant VP based on these elements. What is more, the proposed architecture does not contain fully-connected layers and, thus, less vulnerable to overfitting and transposed FHT layer instead of the direct one allows us to treat answer coordinates normally instead of invoking piece-wise projective transform.

The experiments thus conducted have shown that a trained network with the proposed architecture has a significantly higher quality of detection and at the same time a much lower number of trainable parameters. In addition, due to the absence of fully connected layers, the neural network builds its answer regardless of the position and hence is not overfitted on the positions of correct answers in the training sample.

As part of further studies, it is planned to test this approach on other tasks from image segmentation to determining the orientation of objects. Additional research is needed for the use of a special activation function after the layers of the direct and transposed Hough Transform. It is also planned to use other error functions to improve the learning process and convergence.

*Authors' information*

**Alexander Vladimirovich Sheshkus** (b. 1986) received the B.S. and M.S. degrees in Applied Physics and Mathematics from Moscow Institute of Physics and Technology (State University), Moscow, Russia, in 2009 and 2011, respectively. He is currently the head of machine learning department in Smart Engines, and a researcher in FRC "Computer Science and Control" of RAS. His research interests include deep neural networks, computer vision and projective invariant image segmentation. E-mail: *astdcall@gmail.com* .

**Anastasiya Nikolaevna Chirvonaya** (b. 1998) received the B.S. degree in Applied Mathematics from National University of Science and Technology "MISIS", Moscow, Russia, in 2019. She is currently pursuing the M.S. degree in Applied Science at the same university. She works as a programmer at Smart Engines. Her research interests are computer vision and machine learning. E-mail: *nastyachirvonaya@smartengines.biz* .

**Matveev Daniil Mikhailovich** (b. 1999) is a specialist student in Lomonosov Moscow State University at the Department of Mathematics and Mechanics. He works as a researcher at Smart Engines. His research interest is algorithmic statistics. E-mail: *matveev@smartengines.com*.

**Nikolaev Dmitry Petrovich** (b. 1978) - Ph.D. in Physics and Mathematics, a head of the laboratory at the Institute for Information Transmission Problems of RAS. Graduated from Moscow State University in 2000. Research interests are machine vision, algorithms for fast image processing, pattern recognition. E-mail: *dimonstr@iitp.ru* .

**Arlazarov Vladimir Lvovich** (b. 1939) Dr. Sc., corresponding member of the Russian Academy of Sciences, graduated from Lomonosov Moscow State University in 1961. Currently he works as head of sector 9 at the Institute for Systems Analysis FRC "Computer Science and Control" RAS. Research interests are game theory and pattern recognition. E-mail: *vladimir.arlazarov@gmail.com* .